# AN INEQUALITY PARADIGM FOR PROBABILISTIC KNOWLEDGE

## The Augmented Logic of Conditional Probability Intervals

### Benjamin N. Grosof

Computer Science Department
Stanford University
Stanford, California 94305
(415) 497-4368, 497-2273
*Arpanet: grosof@sushi*

## Abstract and Introduction

An important aspect of knowledge representation in AI systems is how to represent and reason with *probabilistic statements*. We observe that a starting set of probabilistic statements, each assigning a unique value to the probability of some sentence (perhaps conditional on some other sentence), in general does *not* determine a unique value for every sentence of interest. Rather, this *probabilistic theory* only determines an upper and a lower bound on the probability for each sentence of interest. Most approaches to probabilistic reasoning in AI nevertheless have been oriented toward unique-valued probability (e.g. MYCIN [17,1], PROSPECTOR [5,10], Maximum Entropy [11,12,10,3]). In order to *"converge"* to unique values, they have been forced to make powerful additional assumptions (e.g. of independence), and/or to ask delimited questions (i.e. to perform limited inference); often this has led to problems. The assumptions may be unjustified in the domain, or even outright inconsistent, either internally or in presence of other probabilistic information. The limitations on inference may be undesirable. To what extent a particular probabilistic conclusion depends on the assumptions made is often unclear.

We present a paradigm of probabilistic knowledge as a system of inequalities over a space of propositions (augmented by other constraints, i.e. assumptions, e.g. of various (in)dependencies). We introduce the notion of an upper-lower probability distribution, which generalizes both the usual probability distribution notion, and the Belief function of the Dempster-Shafer theory of evidence[16,4]. This provides a unifying framework for:

- representing ignorance and partial information;
- representing the effects of various kinds of (in)dependence assumptions;
- confirmation and evidential combination;
- Dempster-Shafer theory;
- arbitrary deductive inference, including forward and backward chaining, indeed any resolution;
- MYCIN;
- PROSPECTOR; and
- Maximum Entropy;

The inequality paradigm provides a framework to analyze many of the different aspects of, and proposed approaches to, probabilistic knowledge representation and reasoning. Moreover, it suggests how to integrate the various approaches in a coherent manner, taking the best and most apt aspects of each. For example, we can combine evidential reasoning, and the explicit representation of ignorance and partial information (advantages of Dempster-Shafer) with unrestricted rule-based inference using conditional probabilities and Bayes' Rule (an advantage of Bayesian approaches, notably PROSPECTOR). Thus the view of probabilities as bounded intervals offers advantages not only for conceptualization and analysis, but for novel synthesis as well.

# Unconditional Probabilistic Theories

Our initial[1] definition of a *probabilistic statement* is an axiom of the form:

$$p(A) = t$$

which is taken to mean: "the probability of A is t". Here A is an arbitrary sentence, more precisely a closed wff, in some logical language L (we will use first-order logic in our examples). The value t is some real number in the closed interval $[0,1]$[2]. This definition has the axiom state a **unique value** for the **unconditional** probability of A.

(Note that ordinary "non-probabilistic" logical statements (axioms) are just the special case of certainty; i.e. the axiom A in usual logic just corresponds to the probabilistic statement: $(p(A) = 1)$.)

Given an initial set $K = \{(p(A_i) = t_i)\}$ of such probabilistic statements (axioms), we induce (entail) a *probabilistic theory* via *probabilistic logic* (cf. [14]). From the set $S = \{A_i\}$ of sentences appearing in the statements (along with any other sentences whose probability we happen to be interested in), we induce, model-theoretically, a closed space W of propositions over which the probabilistic theory is defined. W is the set of equivalence classes of interpretations for the sentences S, i.e. the set of consistent possible worlds corresponding to truth assignments to $S$[3]. (Here consistency may be relative simply to L, or to some "background" (non-probabilistic) theory in L.) This propositional space W has the structure of a propositional logical language, i.e. it is isomorphic to the power set 2↑F of a set F of "primitive" propositions. Thus W is isomorphic to the notion of a *"frame of discernment"* employed in Dempster-Shafer theory. There F is viewed as a set of distinct "primitive" possibilities[4].

The initial set of statements K *probabilistically entails* ([14]) in general a set of bounded intervals, i.e. inequalities, rather than simply unique values, for the probabilities of the elements of W. That is, we have for each member $W_i$ of W:

$$q_i \leq p(W_i) \leq r_i$$

where $q_i$ and $r_i$ are real numbers in $[0,1]$. We refer to this set of bounds as an *upper-lower probability distribution* since for the probability of each $W_i$ we have an upper and a lower bound, i.e. we can regard the value of $p(W_i)$ as an interval. Since $p(W_i) + p(\neg W_i) = 1$, the upper-lower distribution is just equivalent to the set of lower bounds alone[5], and to the set of upper bounds alone.

The upper-lower probability distribution generalizes both the usual (unique-valued) probability distribution notion, and the Belief function of the Dempster-Shafer theory of evidence. Note that the upper-lower distribution *represents explicitly ignorance and partial information (underdetermination)*. Complete ignorance about the probability of $W_i$ is expressed as an interval $[0,1]$[6].

We would like to regard the upper-lower distribution on W, which we will call P2(W), as a set of concluded (entailed) probabilistic statements (axioms), just like K. This motivates a more general definition of a probabilistic statement as an axiom of the form:

$$q \leq p(A)$$

(or alternatively, as of the forms:

---

[1] We will be treating intervals later on.

[2] However, this is not essential; we might imagine using discretized values; or another range, e.g. [-1,1]; or indeed some arbitrary transform of the usual probability measure which retains its standard axiomatic properties. MYCIN and PROSPECTOR demonstrated the psychological and computational usefulness of alternative, isomorphic representations of probability [5,17,9].

[3] See [14] for more details, including the "semantic tree" method for deriving W from S

[4] and is usually symbolized by $\Theta$

[5] hence Shafer's term "lower probability distribution"

[6] "vacuous" belief in the terminology of Shafer [16]

$$p(A) \leq r \quad ; \text{ or } \quad q \leq p(A) \leq r \quad ).$$

The crux is that we have inequalities and bounded intervals for the probabilities of sentences. Our previous definition is just a special case corresponding to a degenerate interval, i.e. a pair of coinciding bounds.

The more general definition of a probabilistic statement has a nice closure property. If our initial K comprises axioms stating unique values, we in general entail an upper-lower distribution. If our initial K comprises axioms stating bounds or intervals, we in general still entail an upper-lower distribution. Thus starting with bounds we in general end with bounds; starting with equalities (unique values) we in general do no better: we end with bounds rather than simply with equalities.

Another way to view the axioms K is as *constraints* on the possible candidates for the ordinary (single-valued) probability distribution $P(W)$. Thus we can think of the probabilistic theory as a set of constraints as well as as an upper-lower distribution. From the perspective of constraints, we may have two kinds of problems in a probabilistic theory: inconsistency, which can arise in the presence of overdetermination; and underdetermination, i.e. intervals too wide for our purposes. A variety of relaxation techniques have been proposed or employed to cope with inconsistency, involving both automatic and interactive (interviewing the user) aspects [15,11]. Underdetermination can only be overcome by adding more constraints. This means either enlarging K, or adding other sorts of constraints, e.g. various (in)dependence assumptions (we will discuss this more later).

## Conditional Probabilistic Theories

We would like also to consider conditional probabilities, so as to represent probabilistic *rules*, i.e.

```
"if <antecedent>, with certainty, then <consequent>, with <probability interval>".
```

We thus generalize yet again our notion of a probabilistic statement, to be an axiom of the form:

$$q \leq p(A \mid B)$$

where A and B are both sentences in L, and q is a real number in [0,1]. If we view this as a rule, B is the antecedent; and A is the consequent. We will refer to this axiom as a *conditional probabilistic inequality*; we say it is a *CPI*(-form) axiom.

Using the definition of conditional probability, the above CPI axiom is equivalent to:

$$q \times p(B) \leq p(A \wedge B)$$

which is an inequality constraint on unconditional probabilities.

An initial set of axioms K of this more general form again entails a probabilistic theory. Our set of sentences S must include $B_i$ as well as $A_i$ for each axiom in K (equivalently, both $B_i$ and $\{A_i \wedge B_i\}$, as well as $A_i$), in inducing W. And our resultant theory is an upper-lower distribution on not just $(W)$ but rather on $(W \mid W)$, i.e. bounds on all conditional probabilities over the propositional space W. Thus we have generalized the notion of probabilistic logic in [14] to extend to arbitrary conditionals, as well as to initial axioms stating bounds rather than single values.

We can think of unconditional probabilities as a special case of conditional probabilities, where the antecedent is the (trivial) sentence (proposition) "True".

## Additional Assumptions As Constraints

We may wish to constrain our probabilistic theory entailed by K, by the use of additional assumptions not expressed as CPI-form axioms. This has commonly been done in AI, in order to ease the task of probabilistic inference, and/or to "flesh out" underdetermined, incomplete, undesirably-partial probabilistic knowledge. We may or may not want to call these assumptions part of our "probabilistic knowledge" in the sense that K is. Often, they are thought of as only invoked in inference. However, since knowledge representation properly includes the reasoning operations on the formal representational structures (in analogy to abstract data types), they do comprise (a perhaps "auxiliary" sort of) probabilistic knowledge.

An important type of such assumption, employed heavily in PROSPECTOR, is that of *conditional independence*, i.e. an assumption of the form:

"The propositional subspaces C and D are statistically independent, conditional on the propositional subspace G."

$$p(C_i \wedge D_j \mid G_k) = p(C_i \mid G_k) \times p(D_j \mid G_k) \quad \text{for all } C_i \text{ in } C, D_j \text{ in } D, \text{ and } G_k \text{ in } G.$$

We can view Dempster's Rule in Dempster-Shafer theory, as well as the entropy maximization in the Maximum Entropy method, as other types of assumptions not expressed in CPI form.

## Augmented Probabilistic "Logic"

The presence of additional assumptions forces us to adopt a view of probabilistic knowledge which goes beyond probabilisitic logic as discussed above, which we will call "unaugmented" probabilistic logic. More generally what we have is a system of constraints on $P(W \mid W)$. Those in CPI form comprise K; the others we call collectively D. This system (K & D) "entails" (implies) $P2(W \mid W)$, in the sense of determination by a system of inequalities (including equations). However, unlike in unaugmented probabilistic logic, the entailed $P2(W \mid W)$ is not equivalent to, i.e. does not fully represent, the whole of the probabilistic knowledge. In unaugmented probabilistic logic $P2(W \mid W)$ is equivalent to K. Now $P2(W \mid W)$ is implied by, but does not in turn imply all of, (K & D).

We refer to this more general probabilistic knowledge representation formalism as *Augmented Probabilistic "Logic"*, or *A-PL* for short. We will regard the constraints K and D as axioms in our logic[7]. We will be interested in the probabilistic theory T entailed (in the above sense) by these axioms, where T includes $P2(W \mid W)$. CPI axioms are the traditionally explicit notion of probabilistic knowledge. The augmenting assumptions are traditionally implicit, but **there is a duality between the information provided by K and by D: one can substitute for the other.**

Entailment in A-PL addresses the issues of inconsistency and of underdetermined status, i.e. non-degenerate probability intervals. A-PL makes clear what our probabilistic knowledge is, and in particular what our assumptions are. It separates the question of probabilistic inference from the semantics of our probabilistic theory, i.e. probabilistic entailment. We can then ask about the soundness and completeness of our probabilistic inference[8] (more below).

## Evidence and Confirmation

A considerable body of work in AI has been concerned with aggregating measures of confirmatory and disconfirmatory evidence for a common set of propositions. On the face of it, this problem may not appear to be compatible with the usual, classical notion of probability. One apparent difficulty is that if we formalize one piece of evidence as:

$$q_1 \leq p(A) \leq r_1 \quad ; \text{ and another as:} \quad q_2 \leq p(A) \leq r_2$$

then we may have an inconsistency, e.g. if $r_1 < q_2$. Indeed, some researchers have gone so far as to invent new formalisms and methods, claiming classical[9] probability is inadequate [17,16].

However, we can incorporate some of the leading approaches to (probabilistic reasoning oriented towards) accumulated evidence and confirmation into the classical framework and our inequality paradigm, if we treat each piece of evidence as conditional on an evidential source (event). Because A-PL confronts explicitly the issue of ignorance and underdetermination, it avoids the problem of obligatory, yet unavailable, probabilistic "priors".

For example, we can view a *belief function* $B_1$ on a frame of discernment $H$, in Dempster-Shafer theory, as an upper-lower distribution $P2(H \mid E_1)$[10]. Then if we have two such, we can view Dempster's Rule as a constraint (assumption) relating the combined belief function

$$P2(H \mid E_1 \wedge E_2)$$

to the individual belief functions

---

[7] For convenience, we will usually omit henceforth the scare quotes from the terms "logic" and "entailment" in the context of A-PL; however, think of them as implicit.

[8] Of course, as soon as we attack the question of inference, we need to consider justifications and reason maintenance [6].

[9] sometimes referred to as "Bayesian" after its use of Bayes' Rule to combine probabilistic information

[10] More precisely, the belief function corresponds to the lower bound for the probability of each member of $H$; as we mentioned earlier, this is just equivalent to the upper and lower bounds together. Shafer uses the term "plausibility function" for the upper bounds. [16]

$P2(H \mid E_1)$ and $P2(H \mid E_2)$ .

And so on for $E_1, ..., E_n$, i.e. n belief functions, or "pieces" of evidence.

We can think of this as "joint conditioning" since we are generating a new probabilistic (sub-)theory which is conditional on the joint evidence. Thus we represent the accumulation of evidence by augmenting the evidential conditionality of the upper-lower distribution that we are interested in, and by applying some assumptions in order to arrive at an "informative" such P2, i.e. one which is not "vacuous" in the sense of having a bounded interval of [0 , 1] for all sentences in W. Unaugmented probabilistic logic yields a vacuous jointly-conditioned distribution, hence the need for one or more substantive augmenting assumptions as to how to combine evidence from different sources.

The usual, classical, non-"evidential" approach to probabilistic reasoning can be viewed as conditional (trivially) on only one evidential source event. It avoids the issue of joint conditioning.

The Certainty Factors approach to confirmation, pioneered by MYCIN, can also be reformulated in terms of conditional probabilities. Though the MYCIN formalism has some problems, it can be recast without much violation as isomorphic to probabilities, where the combination of certainty factors CF(H,E1) and CF(H,E2) corresponds to something close to the conditional independence assumption made in PROSPECTOR [5], as well as to a special case of Dempster's Rule. See [9,8] for details.

## Analytic Application of the Paradigm: Entailment

A common, and in many repects desirable, state of affairs is for our entailed upper-lower distribution $P2(W \mid W)$ to be representable in a more compact, simpler form than is possible in general: for example, as a single-valued probability distribution $P1(W \mid W)$; or, as a Dempster-Shafer *mass* function. If this is the case, we say that our probabilistic theory *converges* to the simpler representation (i.e. it requires only a special case of our full probabilistic logic.).

In the Maximum Entropy method of probabilistic reasoning, maximization of the entropy of $P(W \mid W)$ is an augmenting assumption (i.e. part of D) in our probabilistic knowledge. It forces the convergence of the entailed CPI statements to single values. In effect, it is one big, global assumption which adapts flexibly to the other constraints. When examining the entailed CPI statements, however, it is not possible to tell whether a probability was "pinned down" (i.e. determined to have a unique value) by the CPI axioms in K, or instead left completely or partially underdetermined by them. Thus we lose "precision" and "justification" information *about* the probabilities. By comparing the converged distribution with the $P2(W \mid W)$ entailed in the absence of the maximum entropy assumption, we can make this distinction; we can tell what is the justification for our entailed probabilistic beliefs; and we can understand more clearly the effect, and the meaning in context, of the maximum entropy assumption.

Dempster-Shafer theory has two essential aspects from the point of view of A-PL entailment. One is Dempster's Rule, which we discussed earlier. The other is the axiomatic structure underlying the mass function representation, and thus the equivalent belief function representation as well. While every belief function is equivalent to an upper-lower distribution, the converse does not hold: *not every upper-lower distribution is (representable as, i.e. equivalent to) a Dempster-Shafer belief function*. Indeed, it is quite easy to generate examples of theories in unaugmented probabilistic logic which are not expressible (fully equivalently) in the Dempster-Shafer formalism. One such K is[11]:

$0.3 \leq p(A \vee B)$ ; $0.4 \leq p(A \vee C)$ ; $0.5 \leq p(B \vee C)$

where A, B, and C are assumed to form the set F (mutually exclusive and exhaustive) of "primitive" propositions ("singletons" in Dempster-Shafer terminology).

Thus the Dempster-Shafer belief function includes unique-valued probability as a proper special case, but is in turn included as a proper special case by interval-valued probability. It occupies a useful middle position in the spectrum of representational richness and simplicity. **This suggests exploiting the Dempster-Shafer representation possibly independently of employing Dempster's Rule**. We might want to use some type of assumptions to ensure convergence of a P2 to a belief function[12]. In a similar spirit, we may want to look for restrictions (and perhaps generalizations) of the Dempster-Shafer representation, along with associated convergence mechanisms. [7] is an interesting example of such an approach.

---

[11] Example due to John Lowrance of SRI International, private communication, mid-1984.

Indeed, (pure) Dempster-Shafer theory is quite limited in its applicability to probabilistic reasoning in AI because it does not represent *rules*, i.e. arbitrary conditional probabilities. The only type of entailment in Dempster-Shafer is by use of Dempster's Rule, i.e. joint conditioning on the evidential sources of multiple belief functions. Once we understand Dempster-Shafer theory within the framework of A-PL, however, we can see formally how to marry it to probabilistic rules, and to perform, for example, forward and backward chaining.

Placing Dempster-Shafer theory in the framework of A-PL focuses the epistemological and decision-theoretic issues involved in applying it. The Dempster-Shafer notions of mass and belief are not (or at least, formally, need not be!) an issue beyond that of the use of probability bounds or intervals. The main semantic question is that of Dempster's Rule. So far as this author is aware, Dempster's Rule has resisted a clear intuitive analysis of its underlying assumptions. The correspondence of Dempster's Rule in the special case of unique-valued probability to something akin to a conditional independence assumption [8] offers some hope that such an understanding will be forthcoming.

Another, more limited example of a commonly-used assumption is the "fuzzy logic" rule of combination used to entail or infer the probability of a conjunction or a disjunction from the probabilities of the conjuncts or disjuncts. This was employed both in PROSPECTOR and in MYCIN. Formally, it says:

$$p(A \wedge B) = \min(p(A), p(B)) \quad ; \quad p(A \vee B) = \max(p(A), p(B))$$

for arbitrary propositions A and B (the conditional form is analogous). This assumption was apparently used because of the perceived need to maintain unique-valued probability. In fact, it corresponds to the assumption of maximal dependency between A and B, i.e. either that $p(A|B) = 1$ or that $p(B|A) = 1$. This is both strong and typically unjustified, as well as unsound in general as a rule of inference; its use in PROSPECTOR led to inconsistencies, and PROSPECTOR was accordingly forced to resort on occasion to renormalizing probabilities. If we allow ourselves to use a bounds representation, however, we can use the sound rule(s) of inference:

$$p(A \wedge B) \leq \min(p(A), p(B)) \quad ; \quad \max(p(A), p(B)) \leq p(A \vee B)$$

Thus if we have $p(A) = 0.3$ and $p(B) = 0.5$ we can use:

$$0 \leq p(A \wedge B) \leq 0.3 \quad ; \text{ and } \quad 0.5 \leq p(A \vee B) \leq 1.$$

Correlation sign information, employed for example in Greg Cooper's NESTOR [3], is also naturally represented as an inequality. Negative correlation is represented as an axiom of the form:

$$p(A \wedge B) \leq p(A) \times p(B) \quad ;$$

positive correlation with the inequality in the other direction. This type of information is often available from qualitative causal knowledge.

### Analytic Application of the Paradigm: Inference

Probabilistic inference is typically directed toward the entailed component $P2(W | W)$ of T. It is a process of closure. We start with a data structure representing directly only the CPI axioms in K, i.e. let $K = \{Ki\}$ where $K_i$ is the axiom in the canonical form:

$$q_i \leq p(A_i | B_i)$$

In the beginning, all other bounds are at the trivial extremes of 0 and 1. By using D as well as the "tautologies" of unaugmented probabilistic logic proper, we proceed to narrow the bounds on probabilities $p(W_i | W_j)$.

Even if we want to converge ultimately to unique-valued probability, it is useful to think about the intermediate states of inference in terms of bounds. One view of probabilistic inference is as constraint propagation. We can generalize the concept of a Local Event Group (LEG) [10] to that of a probabilistic sub-theory in A-PL, with an associated propositional sub-space of $W_i$ as well as a group of axioms, both in CPI form and otherwise.

---

[12] John Lowrance of SRI International proposed using a maximum entropy principle to converge to a mass function, private communication, mid-1984.

Inference consists of applying constraints, but also, in our general scheme, of changing the conditionality on evidence sources. Usually we will not want to explore the whole space $(W | W)$.

Decisions based on probabilistic conclusions are often made by either ranking, i.e. which of several alternatives is more likely or has greatest probabilistically-weighted utility, or by thresholds, i.e. does some alternative have a sufficient probability or probabilistically-weighted utility. Thus bounds will often suffice for decision-making. This may obviate applying, for convergence's sake, strong assumptions which are not justified by our knowledge of the problem domain, and which may even be in conflict (i.e. inconsistent) with that knowledge.

A-PL enables us to take a least-commitment or less-commitment approach to probabilistic reasoning, and allows us to clarify what probabilistic justification constitutes, both in general, and in particular (a problem raised in [6] for example).

Given our newly-developed notions of entailment and inference for probabilistic theories in augmented probabilistic logic, we can hope for fruitful insights into probabilistic reasoning schemes by addressing issues of traditional concern for ordinary non-probabilistic logics: foremost soundness and completeness. We define a set of probabilistic inferences as *sound* iff for each sentence $W_i$ the inferred probability interval for $W_i$ is a superset of the entailed probability interval for $W_i$. Intuitively, this corresponds to the inference being no "stronger" than the entailment, where "stronger" means a tighter bound. We define a set of probabilistic inferences as *complete* iff the inferred probability interval is a subset of the entailed probability interval. Thus *sound and complete* means the inferred distribution is identical to the entailed distribution.

## Conclusion

We have presented a paradigm of probabilistic knowledge in terms of interval bounds, i.e. inequalities, augmented by additional assumptions (about which we have not said much in general), embodied primarily in the augmented logic of interval-valued conditional probability, A-PL. This is a formal, abstract representation, not a practical data structure for implementing probabilistic knowledge representation[13].

The inequality paradigm lets us represent explicitly the incompleteness of our probabilistic knowledge both "logically", i.e. in terms of "entailment", and inferentially. It clarifies the effects of our assumptions, especially those we make to converge to a computationally simpler and more decision-theoretically useful theory. It provides a framework to analyze and combine the representations and assumptions used in different approaches to probabilistic reasoning. Finally, it enables us to integrate the advantages and capabilities of the different approaches, for example evidential reasoning and if-then chaining.

## ACKNOWLEDGEMENTS

This work was supported by the author's National Science Foundation Graduate Fellowship, by the KSL-MRS Project at Stanford University, and by the following grants: NSF MCS 82-06565; NSF MCS 81-04877; DARPA N00039-83-C-0136; ONR N00014-81-K-0004. I thank Peter Cheeseman, Gregory Cooper, my adviser Michael Genesereth, Michael Georgeff, Matthew Ginsberg, Jean Gordon, Kurt Konolige, John Lowrance, and Nils Nilsson for valuable discussions, and for their encouragement and helpfulness.

---

[13] See [14,6] for a discussion of some issues in implementing probabilistic logic with a practical bounds representation. It is worth noting that CPI-form axioms retain linearity in [14]'s scheme:

the statement $\{ p(A | B) \leq r \}$ may be written as the linear inequality $\{ p(A \wedge B) - ( r \times p(B) ) \leq 0 \}$